\documentclass[conference]{IEEEtran}
\usepackage{times}
\pdfoutput=1

\usepackage[numbers]{natbib}
\usepackage{multicol}
\usepackage{algorithm}
\usepackage{algorithmic}
\usepackage{graphicx}
\usepackage{subfig}
\usepackage{amsmath}
\usepackage{amssymb}

\usepackage{color}

\begin{document}

\title{Data-efficient Deep Reinforcement Learning \\for Dexterous Manipulation}
%TODO: should we include the word "Deep" in the title?, e.g. 'Data-efficient Deep Reinforcement Learning for...'
%Obviously a bit pandering, but it might help it get picked up and read on arXiv?

% You will get a Paper-ID when submitting a pdf file to the conference system
\author{Ivaylo Popov, Nicolas Heess, Timothy Lillicrap, Roland Hafner, \\ Gabriel Barth-Maron, Matej Vecerik, Thomas Lampe,  Yuval Tassa, Tom Erez, Martin Riedmiller \\
DeepMind}

%\author{\authorblockN{Michael Shell}
%\authorblockA{School of Electrical and\\Computer Engineering\\
%Georgia Institute of Technology\\
%Atlanta, Georgia 30332--0250\\
%Email: mshell@ece.gatech.edu}
%\and
%\authorblockA{Twentieth Century Fox\\
%\and
%\authorblockN{James Kirk\\ and Montgomery Scott}
%\authorblockA{Starfleet Academy\\
%San Francisco, California 96678-2391\\

% avoiding spaces at the end of the author lines is not a problem with
% conference papers because we don't use \thanks or \IEEEmembership

%Homer Simpson\authorrefmark{2},
%Montgomery Scott\authorrefmark{3} and
%Eldon Tyrell\authorrefmark{4}}
%Telephone: (800) 555--1212, Fax: (888) 555--1212}
%\authorblockA{\authorrefmark{4}Tyrell Inc., 123 Replicant Street, Los Angeles, California 90210--4321}}

\maketitle

\begin{abstract}

%Tim: It's still very rough, but I think all the pieces are there now.

%Tim: I've begun to think that the best way to write the beginning of the intro goes as follows:
%1. talk about the grasping and stacking problem
%2. hand-engineering is hard
%3. deep learning and RL are a way forward.
%But I haven't tried to re-arrange this yet.  I've copied and pasted the Go abstract above.  It's a nice inspiration -- note that it starts with the problem description.

Deep learning and reinforcement learning methods have recently been used to solve a variety of problems in continuous control domains. An obvious application of these techniques is dexterous manipulation tasks in robotics which are difficult to solve using traditional control theory or hand-engineered approaches. One example of such a task is to grasp an object and precisely stack it on another. Solving this difficult and practically relevant problem in the real world is an important long-term goal for the field of robotics. Here we take a step towards this goal by examining the problem in simulation and providing models and techniques aimed at solving it. We introduce two extensions to the Deep Deterministic Policy Gradient algorithm (DDPG), a model-free Q-learning based method, which make it significantly more data-efficient and scalable. Our results show that by making extensive use of off-policy data and replay, it is possible to find control policies that robustly grasp objects and stack them. Further, our results hint that it may soon be feasible to train successful stacking policies by collecting interactions on real robots.

%produced promising results in

% Recent work has shown the promise of deep learning and reinforcement learning for solving continuous control problems. An obvious next step in robotics is to attempt tackle problems which are hard to solve using traditional or hand-tuned approaches. One example of this kind of problem is generalized grasping and stacking objects. Doing this in the real world is clearly the long-term goal. Here we take a step towards this goal by examining the problem in simulation and provide models and techniques for solving it.
% %- If we cannot solve a simplified (i.e. single rather than generic object) version of the problem in %simulation, then this would tell us something quite strong about the chances of success in the real world %with existing methods+algorithms.
% We also introduce two extensions to the DDPG algorithm which make it significantly more data-efficient and scalable.
\end{abstract}

\IEEEpeerreviewmaketitle

\section{Introduction}
\label{sec:introduction}
%Motivation, robotics, core problems. Task setup: multi-step manipulation problem.

Dexterous manipulation is a fundamental challenge in robotics. Researchers have long been seeking a way to enable robots to robustly and flexibly interact with fixed and free objects of different shapes, materials, and surface properties in the context of a broad range of tasks and environmental conditions.
%
%Want robots that robustly and flexible solve complex manipulation problems and operate successfully over a broad range of task variations and initial conditions.
%
Such flexibility is very difficult to achieve with manually designed controllers. The recent resurgence of neural networks and ``deep learning" has inspired hope that these methods will be as effective in the control domain as they are for perception. And indeed, in simulation, recent work has used neural networks to learn solutions to a variety of control problems from scratch (e.g.\ \cite{gu2016continuous,lillicrap2015continuous,schulman2015high,SchulmanLAJM15, heess2015learning,levine2014learning}).

While the flexibility and generality of learning approaches is promising for robotics, these methods typically require a large amount of data that grows with the complexity of the task. What is feasible on a simulated system, where hundreds of millions of control steps are possible \cite{mnih2016asynchronous}, does not necessarily transfer to real robot applications due to unrealistic learning times.
One solution to this problem is to restrict the generality of the controller by incorporating task specific knowledge, e.g.\ in the form of dynamic movement primitives \cite{schaal2006dynamic}, or in the form of strong teaching signals, e.g. kinesthetic teaching of trajectories \cite{Muelling_IJRR_2013}. Recent works have had some success learning flexible neural network policies directly on real robots (e.g.\ \cite{levine2015end,gu2016deep,yahya2016collective}), but tasks as complex as grasping-and-stacking remain daunting.
%While impressive, none of these have examined a task as complex as the grasp and stack problem.
%Although, these results are impressive they remain constrained in the complexity of the tasks that can be solved.

An important issue for the application of learning methods in robotics is to understand how to make the best use of collected data, which can be expensive to obtain, both in terms of time and money. To keep learning times reasonably low even in complex scenarios, it is crucial to find a practical compromise between the generality of the controller and the necessary restrictions of the task setup. This is the gap that we aim to fill in this paper: exploring the potential of a learning approach that keeps prior assumptions low while keeping data consumption in reasonable bounds. Simultaneously, we are interested in approaches that are broadly applicable, robust, and practical.

%TODO: need to be careful here that we don't rais expectations that we then do not satisfy.

%So far, most of the work has, however, been done in simulation. Learning on a real robot poses additional challenges, e.g.\ due to delays and and noise in the hardware, due to safety concerns during exploration, and, in particular, with regards to the limitations in the amount of data that one can collect in a real-world experiment. Traditionally, this has favored learning with carefully designed features and restrictive and specialized movement representations which encode strong prior knowledge about the task (e.g.\ in the form of dynamic movement primitives \cite{schaal2006dynamic}) as well as the use of demonstration data [REF]. Recent works have had some success learning flexible neural network policies directly on real robots (e.g.\ \cite{levine2015end,gu2016deep,yahya2016collective}). While these results are impressive they nevertheless remain constrained in the complexity of the tasks that are being solved and / or the instrumentation of the robot setup they require.

%\cite{gu2016deep,yahya2016collective} further leverage the possibility of parallel training on multiple physical robots simultaneously.
%[REF].
%Some successes also with large scale data collection setups [REF].

% Keep at the top of page 1/column 2
\begin{figure*}[h!]
\includegraphics[width=2.0\columnwidth]{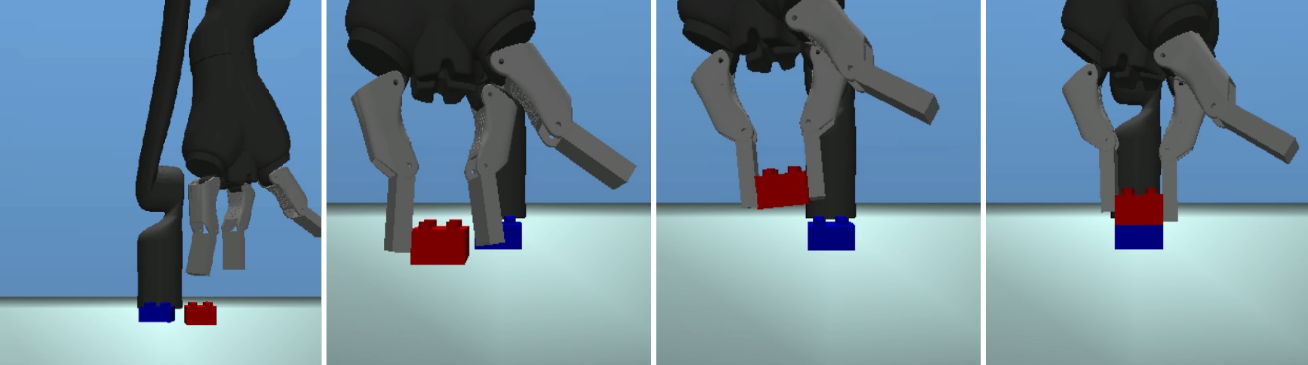}
\caption{Simulation rendering of the Lego task in different completion stages (also corresponding to different subtasks):\\ (a) starting state, (b) reaching, (c) grasping, (also StackInHand starting state) and (d) stacking}
\end{figure*}

%[heess: remove
%A second challenge in addition to the need for data-efficiency is that for complex problems learning from scratch may struggle to find appropriate solutions. This can be mitigated by carefully engineered cost functions, but designing such cost functions can be tedious and error prone process even for tasks of moderate complexity and it is hard to see how it would still be feasible for more complex problems.
%]

%In this paper we study the possibility / methods of learning complex manipulation skills from scratch [What is from scratch?] with minimal prior knowledge.

%In this paper we study the possibility of learning complex manipulation skills with minimal prior knowledge, using a recently developed general purpose model-free deep reinforcement learning algorithm, Deep Deterministic Policy Gradient (DDPG; \cite{lillicrap2015continuous}).
%As a necessary pre-requisite for experiments on a real robot we focus on a feasibility study in a physically realistic simulation.
% TODO: model free hence have very few assumptions; no need for motion primitives etc; eventually no need for instrumentation; GPS relies on demonstration trajectories,  etc.; "learning from scratch"

In this paper we provide a simulation study that investigates the possibility of learning complex manipulation skills end-to-end with a general purpose model-free deep reinforcement learning algorithm.
%\ivo{, minimizing the reliance on specialized movement representations and hand designed features}
The express goal of this work is to assess the feasibility of performing analogous end-to-end learning experiments on real robotics hardware and to provide guidance with respect to the choice of learning algorithm and experimental setup and the performance that we can hope to achieve.
%Our primary consideration in this context is to reduce the amount of environment interaction needed to a level that could be achieved in the real world. We are further aiming to identify broadly applicable, robust, and practical strategies for injecting the prior knowledge that is required for solving a complex manipulation problem from scratch.

%Deep Deterministic Policy Gradient (DDPG; \cite{lillicrap2015continuous}). Our goal is to

%while minimizing the amount of environment interaction. %We are especially interested three questions: is it possible in this setting to achieve the necessary

% TODO: discuss high-dimensionality of action space, long horizon
The task which we consider to this end is that of picking up a Lego brick from the table and stacking it onto a second nearby brick using a robotic arm with 9 degrees of freedom (DoF), six in the arm and three for the fingers in the gripper.
In addition to having a high-dimensional state and action space, the task exemplifies several of the challenges that are encountered in real-world manipulation problems. Firstly, it involves contact-rich interactions between the robotic arm and two freely moving objects. %Furthermore, like many real world manipulation problems it
%is a composite task that involves at least two stages
Secondly it requires mastering several sub-skills (reaching, grasping, and stacking). Each of these sub-skills is challenging in its own right as they require both precision (for instance, successful stacking requires accurate alignment of the two bricks) and as well as robust generalization over a large state space (e.g.\ different initial positions of the bricks and the initial configuration of the arm). Finally, there exist non-trivial and long-ranging dependencies between the solutions for different subtasks: for instance, the ability to successfully stack the brick in the later part of the task depends critically on having picked up the brick in a sensible way beforehand.

%The contribution of this paper is a combination of ideas that not only allow us to solve a complex task successfully and robustly, but also significantly reduce the amount of environment interaction required as well as the overall time to perform an experiment, down to a level where experiments on real hardware seem plausible.

On the algorithm side we build on the Deep Deterministic Policy Gradient (DDPG; \cite{lillicrap2015continuous}), a general purpose model-free reinforcement learning algorithm for continuous action spaces, and extend it in two ways (section \ref{sec:asyncDPG}): firstly, we improve the the data efficiency of the algorithm by scheduling updates of the network parameters independently of interactions with the environment. Secondly, we overcome the computational and experimental bottlenecks of single-machine single-robot learning by introducing a distributed version of DDPG which allows data collection and network training to be spread out over multiple computers and robots.

We further propose two broadly applicable strategies that allow us to inject prior knowledge into the learning process in order to help reliably find solutions to complex tasks and further reduce the amount of environmental interaction. The first of these strategies is a recipe for designing effective shaping rewards for compositional tasks (section \ref{sec:compositeRewards}), while the second (section \ref{sec:startStates}) uses a suitable bias in the distribution of initial states to achieve an effect akin to a curriculum or a form of apprenticeship learning.

In combination these contributions allow us to reliably learn robust policies for the full task from scratch in less than 10 million environment transitions. This corresponds to less than 10 hours of interaction time on 16 robots, thus entering a regime that no longer seems unrealistic with modern experimental setups. In addition, when states from successful trajectories are used as the start states for learning trials the full task can be learned with 1 million transitions (i.e. less than 1 hour of interaction on 16 robots). To our knowledge our results provide the first demonstration of solving complex manipulation problems involving multiple freely moving objects. They are also encouraging as a sensible lower bound for real-world experiments suggesting that it may indeed be possible to learn such non-trivial manipulation skills directly on real robots.

\section{Related work}
\label{sec:related}

%\heess{Arm farm grasping was successful 5-10\% with a random policy - just mention that even our Grasp sub-task is more difficult}

Reinforcement learning approaches solve tasks through repeated interactions with the environment guided by a reward signal that indicates the success or failure of a trial. A wide variety of techniques have been developed that exploit this idea \cite{sutton1998reinforcement}, with a broad distinction often made between value-based and policy search methods. While the former estimate and improve a value function, policy search methods directly optimize the parameters of a policy to maximize cumulative reward. The latter have been routinely applied in robotics, in part because they straightforwardly handle continuous and high-dimensional action spaces \cite{deisenroth2013survey} and applications include manipulation \cite{peters2006policy,kalakrishnan2011learning,pastor2011skill,vonHoof2015learning,levine2015end,gu2016deep,yahya2016collective,gupta2016learning}, locomotion e.g.\ \cite{kohl2004policy,matsubara2006learning}, and a range of other challenges such as helicopter flight \cite{bagnell2001autonomous}.
%Say something about model-based vs. model-free approaches here?

One limitation that has hampered policy search methods is that they can scale poorly with the number of parameters that need to be estimated. This limitation, and other constraints when working with real robotics hardware has led research to focus on the use of manually engineered and restrictive features and movement representations, particularly trajectory-based ones such as spline based dynamic movement primitives. Simplifying the policy space can make learning on real hardware tractable, but it also limits the kinds of problems that can be solved.  In order to solve a problem such as picking up and manipulating an object, more expressive function classes are likely to be needed.

%Furthermore, full expert trajectories are often required to bootstrap learning.
%maybe say something about expert demonstrations being a useful way of injection prior knowledge. Although the focus on full trajectories may be overly restrictive.

The use of rich and flexible function approximators such as neural networks in RL dates back many years, e.g.\ \cite{werbos1990menu,tesauro1995temporal,hunt1992neural,hafner2007neural}. In the last few years there has been a resurgence of interest in end-to-end training of neural networks for challenging control problems, and several algorithms, both value and policy focused have been developed and applied to challenging problems including continuous control, e.g.\ \cite{mnih2015human,mnih2016asynchronous,gu2016muprop,gu2016continuous,lillicrap2015continuous,schulman2015high,SchulmanLAJM15, heess2015learning,levine2014learning}. These methods work well with large neural networks and can learn directly from raw visual input streams. With few exceptions, e.g.\ \cite{hafner2007neural,gu2016deep,levine2015end,yahya2016collective}, they have been considered too data-inefficient for robotics applications.

One exception are guided policy search methods (GPS) \cite{levine2015end,yahya2016collective}. These have recently been applied to several manipulation problems and employ a teacher algorithm to locally optimize trajectories which are then summarized by a neural network policy.
GPS algorithms gain data-efficiency by employing aggressive local policy updates and by performing extensive training of their neural network policy before collecting more real-world data.
The teacher can use model-based \cite{levine2015end} or model-free \cite{yahya2016collective} trajectory optimization. The former can struggle in situations with strong discontinuities in the dynamics, and both rely on access to a well defined and fully observed state space.
%in which trajectory optimization techniques can be performed.
%, which, especially with freely moving objects may require extra instrumentation of the experimental setup. \heess{IS THIS TRUE?}

%Still very rough.
Model-free value function approaches offer an alternative way to handle to the issue of data-efficiency in robotics. Such approaches enable effective reuse of data and do not require full access to the state space or to a model of the environment. One recent work \cite{gu2016deep}, closely related to the ideas followed in this paper, provides a proof of concept demonstration that value-based methods using neural network approximators can be used for robotic manipulation in the real world . This work applied a Q-learning approach \cite{gu2016continuous} to a door opening task in which a robotic arm fitted with an unactuated hook needed to reach to a handle and pull a door to a given angle.  The starting state of the arm and door were fixed across trials and the reward structure was smooth and structured, with one term expressing the distance from the hook to the handle and a second term expressing the distance of the door to the desired angle.  This task was learned in approximately 2 hours across 2 robots pooling their experience into a shared replay buffer.
% This section may need to be re-positioned in the text:

This work thus made use of a complementary solution to the need for large amounts of interaction data: the use of experimental rigs that allow large scale data collection, e.g.\ \cite{pinto2015supersizing}, including the use of several robots from which experience are gathered in parallel \cite{levine2016learning,gu2016deep,yahya2016collective}. This can be combined with single machine or distributed training depending on whether the bottleneck is primarily one of data collection or also one of network training \cite{mnih2016asynchronous}.

Finally, the use of demonstration data has played an important role in robot learning, both as a means to obtain suitable cost functions \cite{boularias2011relative,kalakrishnan2013learning,finn2016guided,gupta2016learning} but also to bootstrap and thus speed up learning. For the latter, kinesthetic teaching is widely used \cite{peters2006policy,kalakrishnan2011learning,pastor2011skill,yahya2016collective}. It integrates naturally with trajectory-based movement representations but the need for a human operator to be able to guide the robot through the full movement can be limiting. Furthermore, when the policy representation is not trajectory based (e.g.\ direct torque control with neural networks) the use of human demonstration trajectories may be less straightforward (e.g.\ since the associated controls are not available). %

%\heess{IS THIS TRUE?? I think we want to say more here, there is a lot of work on bootstrapping / learning from demonstrations in the robotics literature}

%Many of the recent works that attempt to learn skills on real robotics hardware employ either policy search methods or rely on guided policy search. While both approaches have shown good success in practice, the former typically require specialized policy representations with a small number of learnable parameters. The latter work with more general representations such as neural networks but need access to a well defined system state to perform local trajectory optimization using learned or provided dynamics models.
%TODO: this may be restrictive, need state of objects etc.; also not clear how good GPS is with learning from scratch;

%%%%%%%%%%%%%%%%%%%%%%%%%%%%%%%%%%%%%%%%%%%%%%%%%%%%%%%%%%%%%%%%%%%%%%%%
\section{Background}
\label{sec:background}
%Reinforcement learning, Q-learning, deep Q-learning.
In this section we briefly formalize the learning problem, summarize the DDPG algorithm, and explain its relationship to several other Q-function based reinforcement learning (RL) algorithms.

The RL problem consists of an agent interacting with an environment in a sequential manner to maximize the expected sum of rewards. At time $t$ the agent observes the state $x_t$ of the system and produces a control $u_t=\pi(x_t;\theta)$ according to policy $\pi$ with parameters $\theta$. This leads the environment to transition to a new state $x_{t+1}$ according to the dynamics $x_{t+1} \sim p(\cdot | x_t, u_t)$, and the agent receives a reward $r_{t} = r(x_t, u_t)$. The goal is to maximize the expected sum of discounted rewards
$J(\theta)  = \mathbb{E}_{\tau \sim \rho_\theta} \left [ \sum_t \gamma^{t-1} r(x_t, u_t) \right ]$, where $\rho(\theta)$ is the distribution over trajectories $\tau = (x_0, u_0, x_1, u_1, \dots)$ induced by the current policy: $\rho_\theta(\tau) = p(x_0) \prod_{t>0} p(x_t | x_{t-1}, \pi(x_{t-1}; \theta))$.

DPG \cite{silver2014deterministic} is a policy gradient algorithm for continuous action spaces that improves the deterministic policy function $\pi$ via backpropagation of the action-value gradient from a learned approximation to the $Q$-function. %Both the policy and $Q$-function are trained off-policy sampling data from a replay buffer.
Specifically, DPG maintains a parametric approximation $Q(x_t, u_t; \phi)$ to the action value function $Q^\pi(x_t, u_t)$ associated with $\pi$ and $\phi$ is chosen to minimize
\begin{equation}
\mathbb{E}_{(x_t, u_t, x_{t+1}) \sim \bar{\rho}} \left [ ( Q(x_t, u_t; \phi) - y_t)^2 \right ]
\label{eq:dpgQ}
\end{equation}
where $y_t = r(x_t, u_t) + \gamma Q(x_{t+1}, \pi(x_{t+1}))$. $\bar{\rho}$ is usually close to the marginal transition distribution induced by $\pi$ but often not identical. For instance, during learning $u_t$ may be chosen to be a noisy version of $\pi(x_t; \theta)$, e.g. $u_t = \pi(x_t; \theta) + \epsilon$ where $\epsilon \sim \mathcal{N}(0, \sigma^2)$ and $\bar{\rho}$ is then the transition distribution induced by this noisy policy.

The policy parameters $\theta$ are then updated according to
\begin{equation}
\Delta \theta \propto \mathbb{E}_{(x, u) \sim \bar{\rho}} \left [ \frac{\partial}{\partial u}Q(x, u; \phi)\frac{\partial}{\partial \theta}\pi(x; \theta) \right ].
\label{eq:dpgPolicy}
\end{equation}

DDPG \cite{lillicrap2015continuous} is an improvement of the original DPG algorithm adding experience replay and target networks: Experience is collected into a buffer and updates to $\theta$ and $\phi$ (eqs.\ \ref{eq:dpgQ}, \ref{eq:dpgPolicy}) are computed using mini-batch updates with random samples from this buffer. Furthermore, a second set of "target-networks" is maintained with parameters $\theta'$ and $\phi'$. These are used to compute $y_t$ in eqn.\ (\ref{eq:dpgQ}) and their parameters are slowly updated towards the current parameters $\theta$, $\phi$. Both measures significantly improve the stability of DDPG.

%
% Connection to other Q-function based algorithms
%
DDPG bears a relation to several other recent model free RL algorithms: The NAF algorithm \cite{gu2016continuous} which has recently been applied to a real-world robotics problem \cite{gu2016deep} can be viewed as a DDPG variant where the Q-function is quadratic in the action so that the optimal action can be easily recovered directly from the Q-function, making a separate representation of the policy unnecessary. DDPG and especially NAF are the continuous action counterparts of DQN \cite{mnih2015human}, a Q-learning algorithm that recently re-popularized the use of experience replay and target networks to stabilize learning with powerful function approximators such as neural networks. DDPG, NAF, and DQN all interleave mini-batch updates of the Q-function (and the policy for DDPG) with data collection via interaction with the environment. These mini-batch based updates set DDPG and DQN apart from the otherwise closely related NFQ and NFQCA algorithms for discrete and continuous actions respectively. NFQ \cite{riedmiller2005neural} and NFQCA \cite{hafner2011reinforcement} employ the same basic update as DDPG and DQN, however, they are batch algorithms that perform updates less frequently and fully re-fit the Q-function and the policy network after every episode with several hundred iterations of gradient descent with Rprop \cite{riedmiller1993direct} and using full-batch updates with the entire replay buffer. The aggressive training makes NFQCA data efficient, but the full batch updates can become impractical with large networks, large observation spaces, or when the number of training episodes is large. Finally, DPG can be seen as the deterministic limit of a particular instance of the stochastic value gradients (SVG) family \cite{heess2015learning}, which also computes policy gradient via back-propagation of value gradients, but optimizes stochastic policies.
\\
\begin{center}
\begin{tabular}{ccc}
\hline
& Discrete & Continuous \\
\hline
Mini-batch learning\\Target networks & DQN & DDPG, NAF \\
\hline
Full-batch learning with Rprop\\Parameter resetting & NFQ & NFQCA \\
\hline
\end{tabular}
\end{center}

One appealing property of the above family of algorithms is that the use of a Q-function facilitates off-policy learning. This allows decoupling the collection of experience data from the updates of the policy and value networks, a desirable property given that experience is expensive to collect in a robotics setup. In this context, because neural network training is often slow, decoupling allows us to make many parameter update steps per step in the environment, ensuring that the networks are well fit to the data that is currently available.

%%%%%%%%%%%%%%%%%%%%%%%%%%%%%%%%%%%%%%%%%%%%%%%%%%%%%%%%%%%%%%%%%

\section{Task and experimental setup}
\label{sec:setup}
%We used the following experimental setup in this paper. Observations contain
%proprioceptory
%information about the angles and angular velocities of the 6 joints of the arm and 3 fingers of the gripper. In addition, we provide information about the position and orientation of the two bricks and relative distances of the two bricks to the pinch position of the gripper, i.e. roughly the position where the fingertips would meet if the fingers are closed. The 9-dimensional continuous action directly sets the velocities of the arm and finger joints.

The full task that we consider in this paper is to use the arm to pick up one Lego Duplo brick from the table and stack it onto the remaining brick. This "composite" task can be decomposed into several subtasks, including  grasping and stacking. In our experiments we consider the full task as well as the two sub-tasks in isolation as shown in the table below:
%. The following table lists the tasks that we are studying with their respective starting states and reward functions:

%Tim: I switched the ordering of the table.  I think it's nice to build to the full task from bottom to top.  Feel free to change back if there are other arguments.

\begin{center}
\begin{tabular}{ccc}
\hline
& Starting state & Reward \\
\hline
Grasp & Both bricks on table & Brick 1 above table \\
\hline
StackInHand & Brick 1 in gripper & Bricks stacked \\
\hline
Stack & Both bricks on table & Bricks stacked \\
\end{tabular}
\end{center}

In every episode the arm starts in a random configuration with the positioning of gripper and brick appropriate for the task of interest. We implement the experiments in a physically plausible simulation in MuJoCo \cite{todorov2012mujoco} with the simulated arm being closely matched to a real-world Jaco arm\footnote{Jaco is a robotics arm developed by Kinova Robotics} setup in our lab. Episodes are terminated after 150 steps, with each step corresponding to 50ms of physical simulation time. This means that the agent has 7.5 seconds to perform the task. Unless otherwise noted we give a reward of one upon successful completion of the task and zero otherwise.

%We have previously used the same simulation environment for related learning experiments which we then also replicated in an equivalent real-world setup. In these experiments we generally found a good correspondence between learning results in simulation and in reality, especially with respect to the number of environment interactions needed.

The observation vector provided to the agent contains
%proprioceptory
information about the angles and angular velocities of the 6 joints of the arm and 3 fingers of the gripper. In addition, we provide information about the position and orientation of the two bricks and relative distances of the two bricks to the pinch position of the gripper, i.e. roughly the position where the fingertips would meet if the fingers are closed. The 9-dimensional continuous action directly sets the velocities of the arm and finger joints. In experiments not reported in this paper we have tried using an observation vector containing only the raw state of the brick in addition to the arm configuration (i.e. without the vector between the end-effector and brick) and found that this increased the number of environment interactions needed roughly by a factor of two to three.

%Tim: the phrase "of interest" is mysterious here.  We should be more precise.
%After every 30 training episodes the agent is evaluated for 10 episodes.  We used the mean performance at each evaluation phase as the performance measure presented in all plots.  We found empirically that 10 episodes of evaluation gave a reasonable proxy for performance in the studied tasks.

The only hyper-parameter that we optimize for each experimental condition is the learning rate. For each condition we train and measure the performance of 10 agents with different random initial network parameters. After every 30 training episodes the agent is evaluated for 10 episodes.  We used the mean performance at each evaluation phase as the performance measure presented in all plots.  We found empirically that 10 episodes of evaluation gave a reasonable proxy for performance in the studied tasks. In the plots the line shows the mean performance for the set and the shaded regions correspond to the range between the worst and best performing agent in the set. In all plots the x-axis represents the number of environment transitions seen so far at an evaluation point (in millions) and the y-axis represent episode return.

A video of the full setup and examples of policies solving the component and full tasks can be found here: https://www.youtube.com/watch?v=8QnD8ZM0YCo.

%In all experiments for every hyper-parameter configuration of interest we train and measure the performance of a set of agents with different random network parameter initialization. The line shows the mean performance for the set and the shaded regions correspond to the range between the worst and best performing agent in the set. In all plots the x-axis represents the number of environment transitions seen so far at an evaluation point (in millions) and the y-axis represent episode return.

%%%%%%%%%%%%%%%%%%%%%%%%%%%%%%%%%%%%%%%%%%%%%%%%%%%%%%%%%%%%%%%%%

\section{Asynchronous DPG with variable replay steps}
\label{sec:asyncDPG}

%TODO: I think we'll want to lead into this section e.g. in the following way: the replay based Q-learning algorithms are generally pretty data-efficient. However, whether the data efficiency is sufficient for experiments on a real robot where we can at best afford a few thousand episodes remains to be seen. (NFQ/CA are probably pretty data efficient but do not scale). Furthermore, the data efficiency comes at a price: a lot of replay is expensive, which can make the algorithms difficult to use in practice.

%popov: Find this paragraph somewhat confusing: what is experience replay (as in replay memory) and what is experience replay (as in steps).
%Tim: Do you

%As discussed above (section \ref{sec:background}) off-policy learning and storing environment interaction in a memory allows DPG to decouple the collection of environment interactions from the update of the policy and value networks. In this section we study the importance of this property in more detail and find that the effect of under-exploiting it can be dramatic, in some cases making the difference between a task appearing solvable or not. In a second step we then address a caveat of the use of experience replay which is that it can lead to slow learning in wall clock time; this is resolved through the introduction of a distributed asynchronous variant of DPG.

In this section we study two methods for extending the DDPG algorithm and find that they can have significant effect on data and computation efficiency, in some cases making the difference between finding a solution to a task or not.

%Tim: TODO, we should briefly argue that changing the number of minibatch steps is different to turning up the learning rate.  I think that there is a tendency for people to think: "well you should have just turned your learning rate up and you would have obtained similar performance" -- i.e. aren't they roughly equivalent?  This isn't true for 2 reasons: (1) there's only so much we can turn the learning rate up because larger learning rates give rise to divergence & (2) multiple minibatch steps pulls the value back through state space more quickly.  This idea is loosely touched on in the paragraph beginning "One may speculate...", but we should think about addressing it here.
\paragraph{Multiple mini-batch replay steps} Deep neural networks can require many steps of gradient descent to converge. In a supervised learning setting this affects purely computation time. In reinforcement learning, however, neural network training is interleaved with the acquisition of interaction experience, and the nature of the latter is affected by the state of the former -- and vice versa -- so the situation is more complicated. To gain a better understanding of this interaction we modified the original DDPG algorithm as described in \cite{lillicrap2015continuous} to perform a fixed but configurable number of mini-batch updates per step in the environment. In \cite{lillicrap2015continuous} one update was performed after each new interaction step.

\begin{figure}[!h]
\includegraphics[width=1.0\columnwidth]{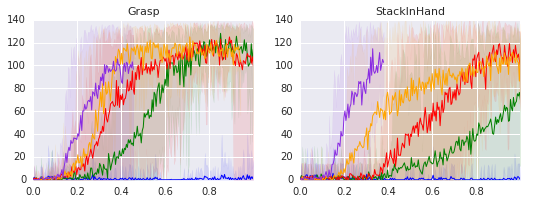}
\caption{Mean episode return as a function of number of transitions seen (in millions) of DPG-R (single worker) on the Grasp (left) and StackInHand (right) task with 1 (blue), 5 (green), 10 (red), 20 (yellow) and 40 (purple) mini-batch updates per environment step \label{fig:DPGR}}
\end{figure}

%Tim: using 5 instead of 'five' is incorrect in terms of proper English prose, but I prefer it for science because when someone's eye scans the page looking for numbers it's much easier to find them this way.
We refer to DDPG with a configurable number of update steps as DPG-R and tested the impact of this modification on the two primitive tasks Grasp and StackInHand. The results are shown in Fig.\ \ref{fig:DPGR}. It is evident that the number of update steps has a dramatic effect on the amount of experience data required for learning successful policies. After one million interactions the original version of DDPG with a single update step (blue traces) appears to have made no progress towards a successful policy for stacking, and only a small number of controllers have learned to grasp. Increasing the number of updates per interaction to 5 greatly improves the results (green traces), and with 40 updates (purple) the first successful policies for stacking and grasping are obtained after 200,000 and 300,000 interactions respectively (corresponding to 1,300 and 2,000 episodes). It is notable that although the improvement is task dependent and the dependence between update steps and convergence is clearly not linear, in both cases we continue to see a reduction in total environment interaction up to 40 update steps, the maximum used in the experiment.

One may speculate as to why changing the number of updates per environment step has such a pronounced effect. One hypothesis is that, loosely speaking and drawing an analogy to supervised learning, insufficient training leads to underfitting of the policy and value network with respect to the already collected training data. Unlike in supervised learning, however, where the dataset is typically fixed, the quality of the policy directly feeds back into the data acquisition process since the policy network is used for exploration, thus affecting the quality the data used in future iterations of network training.

We have observed in various experiments (not listed here) that other aspects of the network architecture and training process can have a similar effect on the extent of underfitting. %like the number of update steps, in further analogy to supervised learning experiments.
Some examples include the type of non-linearities used in the network layers, the size of layers and the learning rate. It is important to note that one cannot replicate the effect of multiple replay steps simply by increasing the learning rate. In practice we find that attempts to do so make training unstable.

\paragraph{Asynchronous DPG}

While increasing the number of update steps relative to the number of environment interactions greatly improves the data efficiency of the algorithm it can also strongly increase the computation time. In the extreme case, in simulation, when the overall run time is dominated by the network updates it may scale linearly with the number of replay steps. In this setting it is desirable to be able to parallelize the update computations.

In a real robotics setup the overall run time is typically dominated by the collection of robot interactions. In this case it is desirable to be able to collect experience from multiple robots simultaneously (e.g.\ as in \cite{yahya2016collective,gu2016deep}). %\heess{[more refs]}). \heess{Maybe also mention that this allows using different robots with slightly different hardware / versions of the task.}

We therefore develop an asynchronous version of DPG that allows parallelization of training and environment interaction by combining multiple instances of an DPG-R actor and critic that each share their network parameters and can be configured to either share or have independent experience replay buffers. This is inspired by the A3C algorithm proposed in \cite{mnih2016asynchronous}, and also analogous to \cite{gu2016deep,yahya2016collective}. We found that this strategy is also an effective way to share parameters for DPG. That is, we employ asynchronous updates whereby each worker has its own copy of the parameters and uses it for computing gradients which are then applied to a shared parameter instance without any synchronization. We use the Adam optimizer \cite{kingma2014adam} with local non-shared first-order statistics and a single shared instance of second-order statistics. The pseudo code of the asynchronous DPG-R is shown in algorithm box \ref{algo:adpgr}.

\begin{algorithm}[h]
  \caption{(A)DPG-R algorithm \label{algo:adpgr}}
  \label{dpgalgo}
  \begin{algorithmic}
    \STATE Initialize global shared critic and actor network parameters:\\
    $\theta^{Q''}$ and $\theta^{\mu''}$
    \STATE {\bf Pseudo code for each learner thread:}
    \STATE Initialize critic network $Q(s, a | \theta^Q)$ and actor
    $\mu(s | \theta^{\mu})$ with weights $\theta^{Q}$ and $\theta^{\mu}$.
    \STATE Initialize target network $Q'$ and $\mu'$ with weights:\\ $\theta^{Q'}
    \leftarrow \theta^{Q}$, $\theta^{\mu'} \leftarrow \theta^{\mu}$
    \STATE Initialize replay buffer $R$
    \FOR{episode = 1, M}
      \STATE Receive initial observation state $s_1$
      \FOR{t = 1, T}
        \STATE Select action $a_t = \mu(s_t | \theta^{\mu}) + \mathcal{N}_t$
        according to the current policy and exploration noise
        \STATE Perform action $a_t$, observe
        reward $r_t$ and new state $s_{t+1}$
        \STATE Store transition $(s_t, a_t,
                r_t, s_{t+1})$ in $R$
        \FOR{update = 1, R}
            \STATE Sample a random minibatch of $N$ transitions
                   $(s_i, a_i,
            r_i, s_{i + 1})$ from $R$
            \STATE Set $ y_i = r_i + \gamma Q'(s_{i + 1},
            \mu'(s_{i+1} | \theta^{\mu'}) | \theta^{Q'}) $

            \STATE Perform asynchronous update of the shared parameters of the critic by minimizing the loss:\\
                   $L = \frac{1}{N} \sum_i (y_i -
                   Q(s_i, a_i | \theta^Q)^2)$
            \STATE Perform asynchronous update of shared parameters of actor policy using the sampled gradient:\\
            \begin{equation*}
                \nabla_{\theta^{\mu''}} \mu|_{s_i} \approx
                \frac{1}{N} \sum_i
                   \nabla_{a} Q(s, a | \theta^Q)|
                   \nabla_{\theta^\mu} \mu(s | \theta^\mu)|_{s_i}
             \end{equation*}
            \STATE Copy the shared parameters to the local ones:\\
             $\theta^{Q}
            \leftarrow \theta^{Q''}$, $\theta^{\mu} \leftarrow \theta^{\mu''}$
            \STATE Every S update steps, update the target networks:\\
              $\theta^{Q'}\leftarrow \theta^{Q}$, $\theta^{\mu'} \leftarrow \theta^{\mu}$
            \ENDFOR
        \ENDFOR
    \ENDFOR
  \end{algorithmic}
\end{algorithm}

%We will compare the performance of different number of replay steps for ADPG-R with 16 workers. In the asynchronous version of the algorithm the effect of multiple replay steps on data efficiency is again significant with up to 8 fold improvement for StackInHand with 20 replay steps instead of one.

\begin{figure}
\includegraphics[width=1\columnwidth]{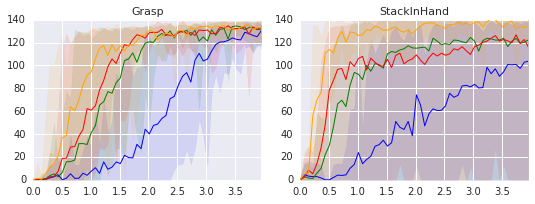}
\caption{Mean episode return as a function of number of transitions seen (in millions) of ADPG-R (16 workers) on the Grasp (left) and StackInHand (right) task. Different colored traces indicate number of replay step as in Fig.\ \ref{fig:DPGR}}
\label{fig:ADPGR}
\end{figure}

Figure \ref{fig:ADPGR} compares the performance of ADPG-R for different number of update steps and 16 workers (all workers performing both data collection and computing updates). Similar to Fig. \ref{fig:DPGR} we find that increasing the ratio of update steps per environment steps improves data efficiency, although the effect appears to be somewhat less pronounced than for DPG-R.

%Figure X compares to the single-worker and asynchronous version of DPG-R. In both cases we chose the best performing number of replay steps and learning rate. As we can see, the use of multiple workers doesn't affect data efficiency for StackInHand and has a slight negative effect for Grasp. This means that the time to reach a certain performance in terms of number of transitions or replay steps per worker is 16x lower for StackInHand and roughly 8x lower for Grasp (bottom row in Figure X). We believe that if a higher number of replay steps is used this small inefficiency will disappear for any task. These results show that scaling up the number of workers for training and environment interaction, i.e. number of robots in our case, can be an extremely effective way in collecting more data and learning policies for more complex dexterous manipulation tasks. We make extensive use of asynchronous DDPG for the rest of the experiments.

\begin{figure}[!h]
\includegraphics[width=1\columnwidth]{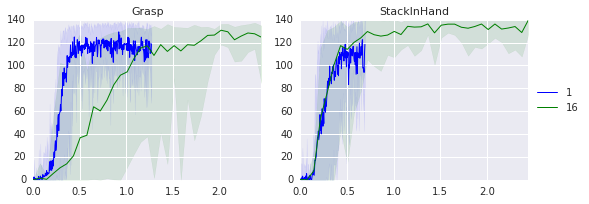}
\includegraphics[width=1\columnwidth]{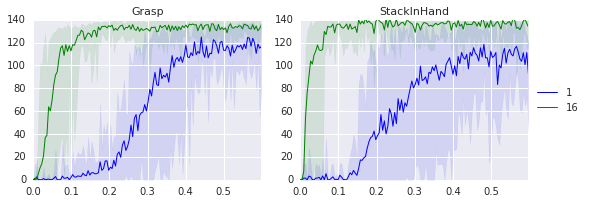}
\caption{Figure with two panels: (a) Grasp; (b) StackInHand; 16 workers vs single worker in data (total for all workers) and "wallclock" (per-worker) time in millions of transitions with best replay step and learning rate selection.
\label{fig:DPGvsADPG}}
\end{figure}

Figure \ref{fig:DPGvsADPG} (top row) directly compares the single-worker and asynchronous version of DPG-R. In both cases we choose the best performing number of replay steps and learning rate. As we can see, the use of multiple workers does not affect overall data efficiency for StackInHand but it reduced roughly in half for Grasp, with the note that the single worker still hasn't quite converged.

Figure \ref{fig:DPGvsADPG} (bottom row) plots the same data but as a function of environment steps \textit{per worker}. This measure corresponds to the optimal wall clock efficiency that we can achieve, under the assumption that communication time between workers is negligible compared to environment interaction and gradient computation (this usually holds up to a certain degree of parallelization). This theoretical wall clock time for running an experiment with 16 workers is about 16x lower for StackInHand and roughly 8x lower for Grasp.

Overall these results show that distributing neural network training and data collection across multiple computers and robots can be an extremely effective way of reducing the overall run time of experiments and thus making it feasible to run more challenging experiments. We make extensive use of asynchronous DPG for remaining the experiments.

%In various experiments with manipulation and toy control task, we have observed that both of the techniques introduced here for DDPG have a very similar effect on other minibatch Q-function based algorithms, such as DQN and NAF. There is indication to believe that discretization and the quadratic advnatage function in NAF do not affect data-efficiency as long as we fit our networks better with different number of replay steps.

%It seems that we can analyze the effect of algorithm choice on data-efficiency more sensible if we take extra care to use the data we acquire: e.g. adding a surrogate loss to single-replay DDPG, e.g. predicting the reward, could indirectly increase the fit of the network for the original task and exhibit signs of improved data-efficiency, which could then partially or completely disappear in another experiment in which both algorithms used a tuned number of replay steps. This realization can be especially useful in transfer learning experiments.

%%%%%%%%%%%%%%%%%%%%%%%%%%%%%%%%%%%%%%%%%%%%%%%%%%%%%%%%%%%%%%%%%

\section{Composite shaping rewards}
\label{sec:compositeRewards}

In the previous section we discussed how the ability of DDPG to exploit information that is available in the acquired interaction data affects learning speed. One important factor that determines what information is available from this data is the nature of the reward function. The reward function in the previous section was "sparse" or "pure" reward where a reward of 1 was given for states that correspond to successful task completion (brick lifted above 3cm for grasp; %\heess{XYZ}
for stack) and 0 otherwise. For this reward to be useful for learning it is of course necessary that the agent is able to enter this goal region in state space with whatever exploration strategy is chosen. This was indeed the case for the two subtasks in isolation, but it is highly unlikely for the full task: without further guidance na\"{i}ve random exploration is very unlikely to lead to a successful grasp and stack as we also experimentally verify in Fig.\ \ref{fig:compositeReward}.

One commonly used solution to this problem is to provide informative shaping rewards that allow a learning signal to be obtained even with simple exploration strategies, e.g.\ by embedding information about the value function in the reward function for every transition acquired from the environment. %, typically in the whole state space of the problem.
%For instance, if we wanted to learn a policy for locomotion with a bipedal creature, we could introduce a shaping reward corresponding to the current forward velocity and sometimes, in addition, other hints, such as a shaping reward for the height of its head from the ground.
For instance, for a simple reaching problem with a robotic arm we could define a shaping reward that takes into account the distance between the end-effector and the target.

While this a convenient way of embedding prior knowledge about the solution and is a widely and successfully used approach for simple problems it comes with several caveats, especially for complex sequential or compositional tasks such as the one we are interested in here.

Firstly, while a suitable shaping reward may be easy to construct for simple problems for more complex composite tasks, such as the one considered in this paper, a suitable reward function is often non-obvious and may require considerable effort and experimentation.
Secondly, and related to the previous point, the use of a shaping reward typically alters the solution to the optimization problem.%A shaping reward may speed up learning by aiding exploration but at the same time it may lead to sub-optimal solutions if the shaping reward is not the real reward function of interest.

% \footnote{
% Although it is possible to construct unbiased shaping rewards \heess{\cite{XYZ}} for a given MDP this requires knowledge of its value function and hence effectively of the solution.
% }

The effect of this can be benign but especially when it comes to complex tasks a small mistake may lead to complete failure of learning as we will demonstrate below. %\heess{maybe mention that this can be hard to distinguish from simply a failure of the RL algorithm to optimize the reward appropriately}
Thirdly, in a robotics setup not all information that would be desirable to define a good shaping reward may be easily available. For instance, in the manipulation problem considered in this paper determining the position of the Lego bricks requires extra instrumentation of the experimental setup.

In this section we propose and analyze several possible reward functions for our full Stack task, aiming to provide a recipe that can be applied to other tasks with similar compositional structure.
Shaping rewards are typically defined based on some notion of distance from or progress towards a goal state. We attempt to transfer this idea to our compositional setup via, what we call, composite (shaping) rewards. These reward functions return an increasing reward as the agent completes components of the full task. They are either piecewise constant or smoothly varying across different regions of the state space that correspond to completed subtasks. In the case of Stack we use the reward components described in table \ref{table:compositeReward}.

\begin{table}
\begin{tabular}{c|p{0.4\columnwidth}|c}
\hline
\multicolumn{3}{c}
{\bf Sparse reward components} \\
\hline
Subtask&Description&Reward \\
\hline
Reach Brick 1 & hypothetical pinch site position of the fingers is in a box around the first brick position &
0.125 \\
\hline
Grasp Brick 1 &
the first brick is located at least 3cm above the table surface, which is only possible if the arm is holding the brick &
0.25 \\
\hline
Stack Brick 1 & bricks stacked & 1.00 \\
\hline
\multicolumn{3}{c}
{\bf Smoothly varying reward components} \\
\hline
Reaching to brick 1&
distance of the pinch site to the first brick - non-linear bounded &
[0, 0.125] \\
\hline
Reaching to stack &
while grasped: distance of the first brick to the stacking site of the second brick - non-linear bounded  &
[0.25, 0.5] \\
\hline
\end{tabular}
\caption{Composite reward function \label{table:compositeReward}}
\end{table}

These reward components can be combined in different ways. We consider three different composite rewards in additional to the original sparse task reward:\\
%{\bf Sparse stacking}: \textit{Stack brick 1} only, i.e. the agent receives a reward of 1 after having completed the full task, 0 otherwise.\heess{This is the original task - I would say we consider 3 composite rewards in addition to the original sparse reward}\\
{\bf Grasp shaping}: \textit{Grasp brick 1} and \textit{Stack brick 1}, i.e. the agent receives a reward of 0.25 when the brick 1 has been grasped and a reward of 1.0 after completion of the full task.\\
{\bf Reach and grasp shaping}: \textit{Reach brick 1}, \textit{Grasp brick 1} and \textit{Stack brick 1}, i.e. the agent receives a reward of 0.125 when being close to brick 1, a reward of 0.25 when brick 1 has been grasped, and a reward of 1.0 after completion of the full task. \\
\textbf{Full composite shaping}: the sparse reward components as before in combination with the distance-based smoothly varying components.

Figure \ref{fig:compositeReward} shows the results of learning with the above reward functions (blue traces). The figure makes clear that learning with the sparse reward only does not succeed for the full task. Introducing an intermediate reward for grasping allows the agent to learn to grasp but learning is very slow. The time to successful grasping can be substantially reduced by giving a distance based reward component for reaching to the first brick, but learning does not progress beyond grasping. Only with an additional intermediate reward component as in continuous reach, grasp, stack the full task can be solved.%\heess{Maybe we want to speculate as to whether learning of the full task could succeed with an additional sparse intermediate reward for reach to brick 2 while grasped.}

Although the above reward functions are specific to the particular task, we expect that the idea of a composite reward function can be applied to many other tasks thus allowing learning for to succeed even for challenging problems. Nevertheless, great care must be taken when defining the reward function. We encountered several unexpected failure cases while designing the reward function components: e.g. reach and grasp components leading to a grasp unsuitable for stacking, agent not stacking the bricks because it will stop receiving the grasping reward before it receives reward for stacking and the agent flips the brick because it gets a grasping reward calculated with the wrong reference point on the brick. We show examples of these in the video: https://www.youtube.com/watch?v=8QnD8ZM0YCo.

\section{Learning from instructive states}
\label{sec:startStates}

In the previous section we have described a strategy for designing effective reward functions for complex compositional tasks which alleviate the burden of exploration. We have also pointed out, however, that designing shaping rewards can be error prone and may rely on privileged information. In this section we describe a different strategy for embedding prior knowledge into the training process and improving exploration that reduces the reliance on carefully designed reward functions.

%\heess{we could lead this paragraph by comparing the figures for primitive tasks and composite tasks and pointing out that stacking is learned very quickly in isolation but much more slowly as part of the full task}
Specifically we propose to let the distribution of states at which the learning agent is initialized at the beginning of an episode reflect the compositional nature of the task: In our case, instead of initializing the agent always at the beginning of the full task with both bricks on the table we can, for instance, choose to initialize the agent occasionally with the brick already in its hand and thus prepared for stacking in the same way as when learning the subtask \textit{StackInHand} in section \ref{sec:asyncDPG}. Trajectories of policies solving the task will have to visit this region of space before stacking the bricks and we can thus think of this initialization strategy as initializing the agent closer to the goal.

More generally, we can choose to initialize episodes with states taken from anywhere along or close to successful trajectories. Suitable states can be either manually defined (as in section \ref{sec:asyncDPG}), or they can be obtained from a human demonstrator or a previously trained agent that can partially solve the task. This can be seen as a form of apprenticeship learning %\heess{[appropriate terminology]}
in which we provide teacher information by influencing the state visitation distribution.

We perform experiments with two alternative methods for generating the starting states. The first one uses manually defined initial states and amounts to the possibility discussed above: we initialize the learning agent in either the original starting states with both bricks located on the table or in states where the first brick is already in the gripper as if the agent just performed a successful grasp and lifted the brick. These two sets of start states correspond to those used in section \ref{sec:asyncDPG}.  %These states with brick in the hand can be produced by mere asset manipulation in the MuJoCo simulation model and this was already used for the sub-experimental setuptackInHand that we looked at in the previous section.

\begin{figure}[!h]
\includegraphics[width=0.49\columnwidth]{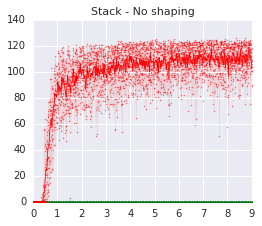}
\includegraphics[width=0.49\columnwidth]{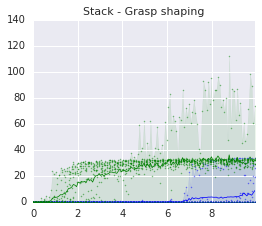}
\includegraphics[width=0.49\columnwidth]{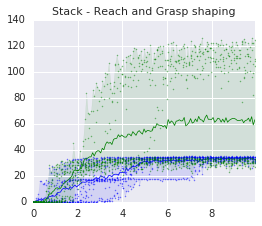}
\includegraphics[width=0.49\columnwidth]{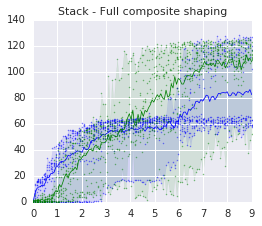}
\caption{Four panels with (a) no progress without extra shaping (b, c, d) different shaping strategies for the composite task with starting states with both bricks on the table (blue), manually defined initial states (green) and initial states continuously on solution trajectories (red). On all plots, x-axis is millions of transitions of total experience and y-axis is mean episode return. Policies with mean return over 100 robustly perform the full Stack from different starting states.
\label{fig:compositeReward}}
\end{figure}

The second method for generating instructive starting states can also be used on a real robot provided a human demonstrator or a pre-trained policy are available. It aims at initializing the learning agent along solution trajectory states in a more fine-grained fashion. We sample a random number of steps for each episode between one and the expected number of steps required to solve the task from the original starting states and then run the demonstrator for this number of steps. The final state of this process is then used as a starting state initialization for the learning agent which then acts in the environment for the remainder of the episode.

The results of these experiments are shown in Figure \ref{fig:compositeReward}. It shows results for the four reward functions considered in the previous section when combined with the simple augmented start state distribution. While there is still no learning for the basic sparse reward case, results obtained with all other reward functions are improved. In particular, even for the second simplest reward function (\textit{Grasp shaping}) we now obtain some controllers that can solve the full task. Learning with the full composite shaping reward is faster and more robust than without the use of instructive states.

The top left plot of Figure \ref{fig:compositeReward} (red trace) shows results for the case where the episode is initialized anywhere along trajectories from a pre-trained controller. We use this start state distribution in combination with the basic sparse reward for the overall case (\textit{Stack without shaping}).
%In this case we of this experiment on the Stack task with sparse reward using ADPG-R with 16 workers and 10 replay steps run on 10 different seeds.
Episodes were configured to be 50 steps, shorter than in the previous experiments, to be better suited to this setup with assisted exploration. During testing we still used episodes with 150 steps as before (so the traces are comparable). We can see a large improvement in performance in comparison to the two-state method variant even in the absence of any shaping rewards. We can learn a robust policy for all seeds within a total of 1 million environment transitions. This corresponds to less than 1 hour of interaction time on 16 simulated robots.

Overall these results suggest that an appropriate start state distribution does not only greatly speed up learning, it also allows simpler reward function to be used. In our final experiment the simplest reward function, only indicating overall experimental success, was sufficient to solve the task. Considering the difficulties that can be associated with designing good shaping rewards this is an encouraging results.

The robustness of the policies that we can train to the starting state variation are also quite encouraging. Table II lists the success rate by task from 1000 trials. You can find a video with trained policies performing the Grasp, StackInHand and Stack tasks from different initial states in the supplementary material.

%\heess{speculate about the two hypotheses: learning both "subtasks" simultaneously; anchoring of the value function}

\begin{table}
\begin{tabular}{p{0.45\columnwidth}|c}
\hline
& Success rate (1000 random starts) \\
\hline
Grasp & 99.2\%  \\
\hline
StackInHand & 98.2\%  \\
\hline
Stack & 95.5\% \\
\hline
\end{tabular}
\caption{Robustness of learned policies. \label{table:robustness}}
\end{table}

%When performing experiments on a real robot there may be constraints as to which initial states can be easily achieved. But in cases where such suitable initializations can be chosen (and we expect that with some creativity when designing  the hardware setup there will be many) this is clearly a promising direction. Furthermore, the approach provides an algorithmically very lightweight way of exploiting demonstration data which also does not rely on demonstrator actions.

%Although this is an impressive results, we have to keep in mind that episode initialization in this experiment can be quite effort and time consuming. Nevertheless, this results gives a nice bound on the amount of experience we need to collect in order to train the neural networks underlying our policy and critic. With the results exploration strategies we seem to require experience which is feasible to collect in a real robotics experiment. We hope that this information will help us design tasks for simulation and real-robot experiments by better assessing their sample complexity.

%%%%%%%%%%%%%%%%%%%%%%%%%%%%%%%%%%%%%%%%%%%%%%%%%%%%%%%%%%%%%%%%%

\section{Conclusion}
\label{sec:conclusion}

%achievements
%limitations
%future work

We have introduced two extensions to the DDPG algorithm which make it a powerful method for learning robust policies for complex continuous control tasks.
%We have particularly focused  on a challenging multi-step dexterous manipulation problem but we expect the extensions to be broadly applicable to other task domains.
Specifically, we have shown that by decoupling the frequency of network updates from the environment interaction we can substantially improve data-efficiency, to a level that in some cases makes the difference between finding a solution or not. The asynchronous version of DDPG which allows data collection and network training to be distributed over several computers and (simulated) robots has provided us with a close to linear speed up in wall-clock time for 16 parallel workers.

In addition, we presented two methods that help to guide the learning process towards good solutions and thus reduce the pressure on exploration strategies and speed up learning. The first, composite rewards, is a recipe for constructing effective reward functions for tasks that consist of a sequence of sub-tasks. The second, instructive starting states, can be seen as a lightweight form of apprenticeship learning that facilitates learning of long horizon tasks even with sparse rewards, a property of many real-world problems. Taken together, the algorithmic changes and exploration shaping strategies have allowed us to learn robust policies for the Stack task within a number of transitions that is feasible to collect in a real-robot system within a few days, or in significantly less time if multiple robots were used for training.

It is of course a challenge to judge the transfer of results in simulation to the real world. We have taken care to design a physically realistic simulation, and in initial experiments, which we have performed both in simulation and on the physical robot, we generally find a good correspondence of performance and learning speed between simulation and real world. This makes us optimistic that our performance numbers also hold when going to the real world. A second caveat of our simulated setup is that it currently uses information about the state of the environment, which although not impossible to obtain on a real robot, may require additional instrumentation of the experimental setup, e.g.\ to determine the position of the two bricks in the work space. To address this second issue we are currently focusing on end-to-end learning directly from raw visual information. Here, we have some first results showing the feasibility of learning policies for grasping with a success rate of about 80\% across different starting conditions.

We view the algorithms and techniques presented here as an important step towards applying versatile deep reinforcement learning methods for real-robot dexterous manipulation with perception.

\bibliographystyle{plainnat}
\bibliography{references}

\newpage
\onecolumn
\appendix

\newcommand{\BZ}{b^{(1)}_z}
\newcommand{\SBX}{s^{B1}_x}
\newcommand{\SBY}{s^{B1}_y}
\newcommand{\SBZ}{s^{B1}_z}
\newcommand{\SB}{s^{B1}}
\newcommand{\SBBX}{s^{B2}_x}
\newcommand{\SBBY}{s^{B2}_y}
\newcommand{\SBBZ}{s^{B2}_z}
\newcommand{\SBB}{s^{B2}}
\newcommand{\SPX}{s^{P}_x}
\newcommand{\SPY}{s^{P}_y}
\newcommand{\SPZ}{s^{P}_z}
\newcommand{\SP}{s^{P}}
\newcommand{\CBB}{C^{(2)}}
\newcommand{\REACH}{\textrm{reach}}
\newcommand{\STACK}{\textrm{stack}}
\newcommand{\GRASP}{\textrm{grasp}}

\subsection{Reward function}

In this section we provide further details regarding the reward functions described in section \ref{sec:compositeRewards}. For our experiments we derived these from the state vector of the simulation, but they could also be obtained through instrumentation in hardware.  The reward functions are defined in terms of the following quantities:
\begin{itemize}
\item $\BZ$: height of brick 1 above table
\item $s^{B1}_{\{x,y,z\}}$: x,y,z positions of site located roughly in the center of brick 1
\item $s^{B2}_{\{x,y,z\}}$: x,y,z positions of site located just above brick 2, at the position where $\SB$ will be located when brick 1 is stacked on top of brick 2.
\item $s^P_{\{x,y,z\}}$: x,y,z positions of the pinch site of the hand -- roughly the position where the fingertips would meet if the fingers are closed..
%\item $s^{B2}_{x},s^{B2}_{y},s^{B2}_{z}$: x,y,z position of a site located roughly in the center of brick 2
%\item $s^{P}_{x,y,z}$: x,y,z position of the pinch site
\end{itemize}

\subsubsection{Sparse reward components}
Using the above we can define the following conditions for the successful completion of subtasks:

\paragraph{Reach Brick 1} The pinch site of the fingers is within a virtual box around the first brick position.
\begin{align}
\REACH =& (|\SBX - \SPX | < \Delta^\mathrm{reach}_x) \land (|\SBY - \SPY | < \Delta^\mathrm{reach}_y) \land  (|\SBZ - \SPZ | < \Delta^\mathrm{reach}_z) \nonumber,
\end{align}
where $\Delta^{\mathrm{reach}}_{\{x,y,z\}}$ denote the half-lengths of the sides of the virtual box for reaching.

\paragraph{Grasp Brick 1} Brick 1 is located above the table surface by a threshold, $\theta$, that is possible only if the arm is the brick has been lifted.
\begin{align}
\GRASP =& \BZ > \theta \nonumber
\end{align}

\paragraph{Stack} Brick 1 is stacked on brick 2. This is expressed as a box constraint on the displacement between brick 1 and brick 2 measured in the coordinate system of brick 2.
\begin{align}
\STACK =& (|\CBB_x (\SB - \SBB)| < \Delta^\mathrm{stack}_x) \land (|\CBB_y (\SB - \SBB)| < \Delta^\mathrm{stack}_y) \land (|\CBB_z (\SB - \SBB)| < \Delta^\mathrm{stack}_z) \nonumber,
\end{align}
where $\Delta^{\mathrm{stack}}_{\{x,y,z\}}$ denote the half-lengths of the sides of the virtual box for stacking, and $\CBB$ is the rotation matrix that projects a vector into the coordinate system of brick 2. This projection into the coordinate system of brick 2 is necessary since brick 2 is allowed to move freely. It ensures that the box constraint is considered relative to the pose of brick 2. While this criterion for a successful stack is quite complicated to express in terms of sites, it could be easily implemented in hardware e.g.\ via a contact sensor attached to brick 2.

\subsubsection{Shaping components}
The full composite reward also includes two distance based shaping components that guide the hand to the brick 1 and then brick 1 to brick 2. These could be approximate and would be relatively simple to implement with a hardware visual system that can only roughly identify the centroid of an object. The shaping components of the reward are given as follows:

\paragraph{Reaching to brick 1}:
\begin{align}
r_{S1}(\SB,\SP) &= 1-\tanh^2(w_1 \| \SB-\SP \|_2) \nonumber
\end{align}
\paragraph{Reaching to brick 2 for stacking}
\begin{align}
r_{S2}(\SB,\SBB) &= 1-\tanh^2(w_2 \| \SB-\SBB \|_2). \nonumber
\end{align}

\subsubsection{Full reward}
Using the above components the reward functions from section \ref{sec:compositeRewards}: \emph{Stack}, \emph{Grasp shaping}, \emph{Reach and grasp shaping}, and \emph{Full composite shaping}  can be expressed as in equations (\ref{eq:rewardStack}, \ref{eq:rewardGraspShape}, \ref{eq:rewardReachGraspShape}, \ref{eq:rewardFullShape}) below. These make use of the predicates above to determine whether which subtasks have been completed and return a reward accordingly.
\begin{align}
r(\BZ, \SP, \SB, \SBB) &= \begin{cases}
1 &\text{if}~ \STACK(\BZ, \SP, \SB, \SBB)\\    0 & \text{otherwise}
\end{cases} \label{eq:rewardStack} \\
r(\BZ, \SP, \SB, \SBB) &= \begin{cases}
1 &\text{if}~ \STACK(\BZ, \SP, \SB, \SBB)\\
0.25 &\text{if}~ \neg \STACK(\BZ, \SP, \SB, \SBB) \land \GRASP(\BZ, \SP, \SB, \SBB)\\
0 & \text{otherwise}
\end{cases} \label{eq:rewardGraspShape}\\
r(\BZ, \SP, \SB, \SBB) &= \begin{cases}
1 &\text{if}~ \STACK(\BZ, \SP, \SB, \SBB)\\
0.25 &\text{if}~ \neg \STACK(\BZ, \SP, \SB, \SBB) \land \GRASP(\BZ, \SP, \SB, \SBB)\\
0.125 &\text{if}~ \neg ( \STACK(\BZ, \SP, \SB, \SBB)  \lor \GRASP(\BZ, \SP, \SB, \SBB) ) \land \REACH (\BZ, \SP, \SB, \SBB) \\
0 & \text{otherwise}
\end{cases} \label{eq:rewardReachGraspShape}\\
r(\BZ, \SP, \SB, \SBB) &= \begin{cases}
1 &\mathrm{if}~ \STACK(\BZ, \SP, \SB, \SBB)\\
0.25 + 0.25 r_{S2}(\SB,\SP) & \mathrm{if}~ \neg \STACK(\BZ, \SP, \SB, \SBB) \land \GRASP(\BZ, \SP, \SB, \SBB)\\
0.125 &\mathrm{if}~ \neg ( \STACK (\BZ, \SP, \SB, \SBB)  \lor \GRASP(\BZ, \SP, \SB, \SBB) ) \land \REACH(\BZ, \SP, \SB, \SBB)\\
0 + 0.125 r_{S1}(\SB,\SP) & \text{otherwise}
\end{cases} \label{eq:rewardFullShape}
\end{align}

\end{document}